\documentclass[sigconf]{acmart}
\usepackage{graphicx}
\usepackage{hyperref}
\usepackage{listings}
\usepackage{booktabs}
\usepackage{multirow}
\usepackage{amsmath}
\usepackage[most]{tcolorbox}
\usepackage{xcolor}
\usepackage{booktabs} 
\usepackage{amsmath}
\usepackage{amsthm}
\usepackage{multirow,tabularx,ragged2e}
\usepackage{adjustbox}
\usepackage{tabularx}
\usepackage{float}
\usepackage{algorithm}
\usepackage{algpseudocode}
\usepackage{makecell}
\usepackage{tcolorbox} 
\usepackage{xcolor}
\usepackage{lipsum}
\usepackage{subcaption}

\setcopyright{acmlicensed}
\copyrightyear{2026}
\acmYear{2026}
\acmDOI{XXXXXXX.XXXXXXX}
\acmConference[SIGSPATIAL '26]{The 34th ACM International Conference on Advances in Geographic Information Systems}{Nov 03--06,
  2026}{Riverside, CA}
\acmISBN{978-1-4503-XXXX-X/2018/06}

\newcolumntype{Y}{>{\raggedright\arraybackslash}X}

\newtcolorbox{highlightbox}{
  colback=yellow!20,   
  colframe=black,      
  boxrule=0.5pt,       
  arc=0pt,             
  left=5pt, right=5pt, 
  top=5pt, bottom=5pt  
}

\copyrightyear{2025}
\acmYear{2025}
\setcopyright{rightsretained}

\begin{document}

\title{Staypoint Detection from Noisy Trajectory Data [Experiment Paper]}
\author{Lance Kennedy}
\orcid{0009-0004-6815-2219}
\affiliation{%
  \institution{Emory University, USA}
  \city{}
  \state{}
  \country{}
}
\email{lance.kennedy@emory.edu}

\author{Hossein Amiri}
\orcid{0000-0003-0926-7679}
\affiliation{%
  \institution{Emory University, USA}
  \city{}
  \state{}
  \country{}
}
\email{hossein.amiri@emory.edu}

\author{Yueyang Liu}
\orcid{}
\affiliation{%
  \institution{Emory University, USA}
  \city{}
  \state{}
  \country{}
}
\email{yueyang.liu@emory.edu}

\author{Riyang Bao}
\orcid{0000-0002-3763-4539}
\affiliation{%
  \institution{Emory University, USA}
  \city{}
  \state{}
  \country{}
}
\email{riyang.bao@emory.edu}

\author{Hanqi Chen}
\orcid{0009-0007-6124-7583}
\affiliation{%
  \institution{Emory University, USA}
  \city{}
  \state{}
  \country{}
}
\email{hanqi.chen@emory.edu}

\author{Mohammad Hashemi}
\orcid{0009-0005-1608-7213}
\affiliation{%
  \institution{Emory University, USA}
  \city{}
  \state{}
  \country{}
}
\email{mohammad.hashemi@emory.edu}

\author{Ruochen Kong}
\orcid{0009-0006-0329-8019}
\affiliation{%
  \institution{Emory University, USA}
  \city{}
  \state{}
  \country{}
}
\email{ruochen.kong@emory.edu}

\author{Xiaotong Liu}
\orcid{0009-0007-2617-8447}
\affiliation{%
  \institution{Emory University, USA}
  \city{}
  \state{}
  \country{}
}
\email{steven.liu@emory.edu}

%

\author{Joon-Seok Kim}
\orcid{}
\affiliation{%
  \institution{Emory University, USA}
  \city{}
  \state{}
  \country{}
}
\email{joonseok.kim@emory.edu}

\author{Shengpu Tang}
\orcid{0000-0002-4213-2015}
\affiliation{%
  \institution{Emory University, USA}
  \city{}
  \state{}
  \country{}
}
\email{shengpu.tang@emory.edu}

\author{Liang Zhao}
\orcid{}
\affiliation{%
  \institution{Emory University, USA}
  \city{}
  \state{}
  \country{}
}
\email{liang.zhao@emory.edu}

\author{Andreas Z{\"u}fle}
\orcid{}
\affiliation{%
  \institution{Emory University, USA}
  \city{}
  \state{}
  \country{}
}
\email{azufle@emory.edu}
\renewcommand{\shortauthors}{}
\begin{abstract}
    Detecting staypoints from raw trajectory data is fundamental to numerous spatial computing applications. This process transforms raw numeric sequences of geolocations into semantically meaningful locations, such as homes, workplaces, or restaurants. Despite its importance for semantic trajectory analysis, staypoint detection lacks standard benchmarks, and existing algorithms have never been systematically evaluated. This gap persists because no publicly available datasets provide both raw individual trajectories and ground-truth staypoint annotations. This benchmark paper addresses this limitation with two key contributions: (1) we introduce 16 large-scale simulated datasets capturing thousands of agents with annotated staypoints across varying trajectory noise levels, and (2) we evaluate nine staypoint detection algorithms—including both state-of-the-art and novel methods—to analyze their robustness to noise. Our evaluation reveals that existing state-of-the-art algorithms perform poorly under realistic noise conditions. Conversely, our proposed unsupervised methods yield substantial improvements, while supervised approaches drastically outperform existing baselines. While these results are very promising, these datasets and methods are only meant as starting points for future research in staypoint detection.

\end{abstract}



\maketitle

\vspace{-0.5cm}
\section{Introduction}
\label{sec:introduction}
Trajectory data captures the location of individual users or objects over time. For example, think of your own trajectory on the last weekend: Where did you go? Where did you stay? But trajectory information only captures the ``Where?'' and the "When?'' of mobility, but not the ``What?'' and ``Why?'': As trajectory data only captures time-stamped geolocations (lat, long), it cannot be directly used to answer questions such as ``What kind of place did you stay at last Saturday afternoon?'' and ``Why did you go to that place?''. Adding such semantic information, that describes the type of a place (home, work, restaurant, recreation, etc.) visited by a trajectory or the activity performed at a place (sleep, work, eat, drink, etc.) leads to the notion of a \emph{semantic trajectory}~\cite{parent2013semantic,ying2011semantic}. Semantic trajectories enable understanding of trajectories beyond two-dimensional time-stamped location signals (when? and where?), but allow to explain behavior (what?) and detect anomaluos behavior (why?). For example, a (raw) trajectory may show a child located far away from school during school hours. A semantic trajectory may explain this deviation by visiting a National History Museum as a class excursion or by visiting a game arcade and skipping school. The former is explainable normal behavior whereas the latter may need to be reported to the parents as an anomaly. Making such distinctions is impossible without semantic information. But mapping raw trajectories to semantic places is not trivial for multiple reasons: 
\begin{itemize}
    \item Location signals are noisy. For example, trajectories collected using GPS are subject to receiver noise (hardware thermal noise), environmental noise (such as signal-reflecting obstacles), and atmospheric/solar interference.
    \item Trajectories are incomplete, as signals may be lost and measuring devices may run out of power or be turned off. 
    \item Locations are ambiguous. For example, a high-rise building may have different shops on different floors, making it challenging to map a location to a unique semantic location.
\end{itemize}

Successfully mapping uncertain raw trajectories to semantic places is critical to unlocking the immense value of location-based data for personal and commercial applications. Over fifteen years ago, the McKinsey Global Institute's "Big Data" report projected a ``\$600 billion annual consumer surplus from global personal location data usage''~\cite{manyika2011big}. Despite this enduring potential, fully capitalizing on this promise remains an open challenge for the spatial computing community. A primary bottleneck hindering progress is the critical shortage of publicly available datasets that concurrently capture high-fidelity location trajectories and rich semantic annotations.

\begin{highlightbox}
Gap 1: As of June 2026, there exists no publicly available dataset that captures both (1) individual-level trajectories and (2) the corresponding semantic staypoints.
\end{highlightbox}
The above claim may sound bold and the interested reader may ask: ``What about GeoLife? What about FourSquare?''. But trajectory datasets like GeoLife~\cite{geolife} capture no ground truth information on staypoints. GeoLife captures trajectories of 182 Users in Beijing, China, but it is not possible to infer the visited places from raw trajectories, as many different places (restaurants, bars, apartments, workplaces) may be physically located in the same building and indistinguishable using two-dimensional coordinates, especially with large urban GPS noise convoluting neighboring buildings and blocks. There exist related works that apply staypoint detection and geocoding to GeoLife to enrich GeoLife data with semantic information~\cite{amiri2023massive}. However, such approaches use simple nearest neighbor matching using building data from OpenStreetMap and it remains unclear to what degree this semantic information is correct as there are no ground truth staypoints for GeoLife to validate the semantic information. Vehicle tracking dataset such as the T-Drive dataset~\cite{yuan2010t}, which contains one-week trajectories of 10,357 taxis in Beijing, China has similar issues: while it does contain individual-level trajectories for these taxis, it does not contain annotated staypoints of passengers.

There are also \emph{Check-In} datasets such as FourSquare Data~\cite{yang2014modeling}, GoWalla Data~\cite{cho2011friendship}, and BrightKite~{Data}~\cite{cho2011friendship}. Such data captures semantically enriched staypoints (2), but without trajectories (1): These datasets capture staypoints using location-based apps that allow uses to notify the app of staying at a place (called a ``Check-In'') such as a coffee shop to receive a discount or a free beverage. While such data captures ground truth annotated staypoints by having users validate their visit to a place, this data does not include any trajectory information outside of the check-ins. Thus, there is no information where the user was outside of a checkin. 
In summary, trajectory datasets provide movement observations but little semantic ground truth, while check-in datasets provide semantics but little continuous mobility information. The lack of publicly available dataset that capture both lead to the second gap address in this work:
\begin{highlightbox}
Gap 2: As of May 2026, there exists no published experimentally validated algorithm for staypoint detection.
\end{highlightbox}
While this statement may also sound surprising, but it is a direct consequence of the lack of a dataset. Without any ground truth data that captures both raw trajectories and true staypoints, it is impossible to validate a staypoint detection algorithm to understand how well these actually work. The most commonly state-of-the-art algorithm for staypoint detection is the Sliding Window Staypoint Detection Algorithm~\cite{li2008mining} published at SIGSPATIAL'08. This algorithm has been implemented in the widely used Trackintel Python library~\cite{Martin_2023_trackintel} commonly used for mobility data analysis. However, the goal of the paper that this algorithm was published in (\cite{li2008mining}) was not staypoint detection, and staypoint detection was only an intermediate step towards the goal of this paper of measuring semantic similarity between users. Due to this, the Sliding Window Staypoint Detection has not been experimentally evaluated such that it remains unclear whether this algorithm is truly able to correctly detect staypoints, especially in the presence of noise. 

The goal of this paper is to close the aforementioned two gaps. To close Gap 1, we provide a benchmark dataset that aims at providing realistic simulated trajectory data, that captures (a) staypoints visited by simulated agents to satisfy their needs, (b) realistic trajectories to move between staypoints in an urban environment, and (c) realistic noise to simulate GPS uncertainty and loss of connection to benchmark staypoint detection algorithms. We provide 16 datasets of a city simulation of Atlanta, GA, USA, at different levels of simulated noise that we hope the community may use to test their own strategies and algorithms for staypoint detection and other tasks. To close Gap 2, we evaluate the existing state-of-the-art~\cite{li2008mining} as well as eight novel algorithms for staypoint detection to see what strategies are best capable for detecting staypoints accurately at different levels of noise. But we note that these approaches are not meant as an exhaustive list of possible staypoint detection algorithms. Instead, this paper is meant as a starting point towards a new research field of staypoint detections using our provided datasets and algorithms as starting points and baselines for the community to find better staypoints detection solutions.



This paper is organized as follows: In section~\ref{sec:related_works}, we review related literature. In Section~\ref{sec:data}, we describe the datasets used in our experiments. In Section~\ref{sec:methodology}, we present a brief description of each staypoint detection algorithm assessed in this paper. In Section~\ref{sec:results}, we present our experimental results, and conclude in Section~\ref{sec:conclusion}.

\vspace{-0.2cm}
\section{Related Works}
\label{sec:related_works}
\vspace{-0.1cm}
As GPS technology becomes more readily available, tools to utilize trajectory data become more important. Trajectory data has many use cases, such as classification~\cite{bian2019trajectory}, modality detection~\cite{zheng2008understanding}, and anomaly detection~\cite{liu2024neural,zhang2024large,zhan2026self}. 
Because raw trajectory data is so large, all of these tasks rely on generating accurate summary data from the raw trajectories, most commonly in the form of staypoints. Some studies highlight the concerns surrounding how data dropout affects the quality of the summary data, noting that real-world data sees large amounts of data dropout which must be handled~\cite{shaukat2016robust}. This could occur for any number of reasons: user error, device malfunction, or dead zones such as tunnels where data transmission is impossible.

Despite this, there is very little precedence for validating the quality of staypoints mined from trajectory data. This is primarily due to a lack of ground-truth: there are many datasets of trajectory data, such as the Geolife GPS trajectory dataset~\cite{zheng2010geolife}, consisting of GPS traces from hundreds of users over multiple transportation modes, or the T-drive trajectory dataset~\cite{yuan2011driving, yuan2010t}, which records GPS data for taxis. However, without users manually recording exactly where and when they made stops during the recorded period, we cannot know this data with any certainty. The users recorded in these datasets experience the GPS noise and data dropout which make mining staypoints a difficult problem in the first place. On the other hand, there are check-in datasets which consist of location-time pairs where users ``check in" to a location, so we know exactly when and where they are, but do not include trajectories to accompany this data due to privacy concerns~\cite{cho2011friendship}.

Several publicly available real-world location-based social network (LBSN) check-in datasets have been widely used in prior research, including Gowalla, BrightKite, Foursquare, and Yelp~\cite{liu2013personalized, cho2011friendship,yang2019revisiting,yelp_open_dataset}. These datasets summarize user check-in behavior by recording interactions between users and physical locations over time, often together with social links or additional contextual information. Gowalla contains about 319K users, 2.8M locations, 36M check-ins, and 4.4M social links over 20 months, making it one of the richest public LBSN datasets with social network information. BrightKite is a smaller but similar dataset, with 58K users, 971K locations, 4.49M check-ins, and 214K links over 30 months. Foursquare is larger in terms of scale, containing 2.7M users, 11.1M locations, and 90M check-ins over five months, but it does not provide social network links. Yelp includes 1M users, 144K locations, and 4.10M check-ins over 36 months, along with useful information such as ratings, reviews, user profiles, and location attributes, but it also lacks social network information. These datasets are important benchmarks for LBSN studies, but they remain limited by sparsity, missing ground truth, privacy concerns, and incomplete social or temporal information~\cite{kim2020location} .

Still, several algorithms exist for mining staypoints from trajectory data, most notably the Sliding Window Staypoint Detection Algorithm~\cite{li2008mining}. In simple terms, this algorithm checks whether successive trajectory points remain within given temporal and spatial thresholds, and if so, identifies a staypoint centered on those points.
Sliding windows were originally introduced as a step in measuring user similarity based on location history~\cite{li2008mining}. This intermediate GPS-processing step has also been used in~\cite{liu2024neural,zhang2024large,zhang2024transferable,zhan2026self}, where the authors used raw GPS trajectories from the GeoLife dataset~\cite{zheng2010geolife} to detect anomalous behavior among users. They selected users with more than 100 staypoints and applied different outlier detection approaches to the resulting data. Their results show high accuracy and AUC, indicating that staypoints are effective features for detecting anomalies in trajectory data. However, there is no ground truth available to verify whether the staypoints extracted from GeoLife are valid and accurate.

The sliding-window algorithm has been widely adopted for staypoint detection, including in notable mobility-analysis tools such as Trackintel~\cite{martin2023trackintel}, MovingPandas~\cite{graser2019movingpandas}, and TransBigData~\cite{yu2022transbigdata}. Some studies have proposed clustering-based improvements to the algorithm~\cite{yuan2011find,yuan2012t}, while other approaches replace the sliding-window procedure with clustering altogether. One such example is ST-DBSCAN~\cite{BIRANT2007208}, which extends the density-based clustering algorithm DBSCAN to mine staypoints from trajectory data. In this experiment, benchmark, and experience paper, we evaluate existing approaches and develop new methods to provide a comprehensive benchmark of staypoint detection algorithms under varying levels of noise and data quality.

\vspace{-0.2cm}
\section{Description of Data}
\label{sec:data}
\vspace{-0.1cm}
The primary dataset used for these experiments is a simulated trajectory dataset based in Atlanta, GA, USA. The trajectories are created using the Patterns of Life (PoL) simulation~\cite{amiri2024patterns, kim2020location}. It consists of 500 agents simulated for a period of 30 days with a sampling rate of 1 minute per observation. Each observation includes a timestamp, latitude, and longitude for a given agent. 
The PoL simulation uses OpenStreetMap to initialize four types of locations: homes, workplaces, restaurants, and recreational sites. Agents in the simulation mimic real-life mobility patterns by making movement decisions based on realistic needs and behaviors. As a result, agents tend to have consistent home and work locations, while also occasionally visiting restaurants and recreational sites.
As the simulation runs, agents attempt to satisfy their needs by moving between different places. To support realistic movement, the agents use a routing module that determines the shortest paths between locations on a road network. The simulation records agents’ activities and states and saves them in different log files.
One of the most important outputs of the simulation is the agent trajectory log, which records each agent’s location at every tick of the simulation. This trajectory data makes it possible to analyze movement patterns, visited locations, travel behavior, and changes in agent activity over time.

\begin{figure*}
    \centering
    \includegraphics[width=\textwidth]{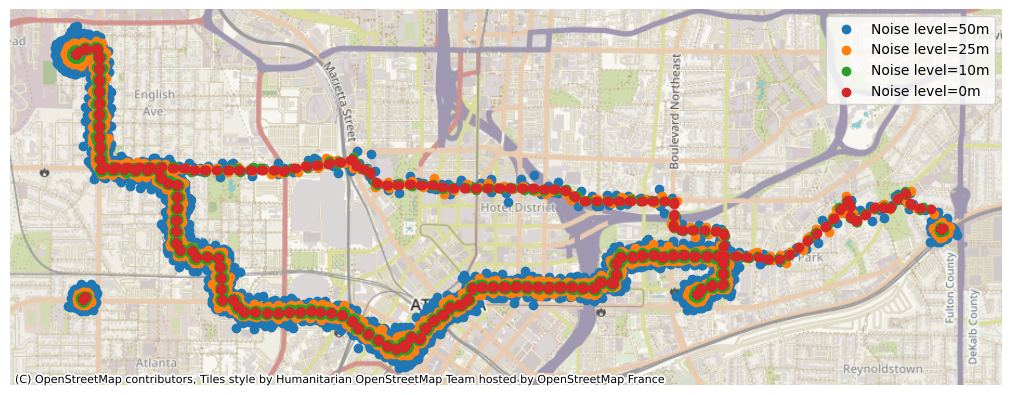}\vspace{-0.5cm}
    \caption{Full trajectory of Agent 0 under different noise levels.}
    \label{fig:traj}
\end{figure*}

\begin{figure}
    \centering
\includegraphics[width=0.9\linewidth,height=6cm]{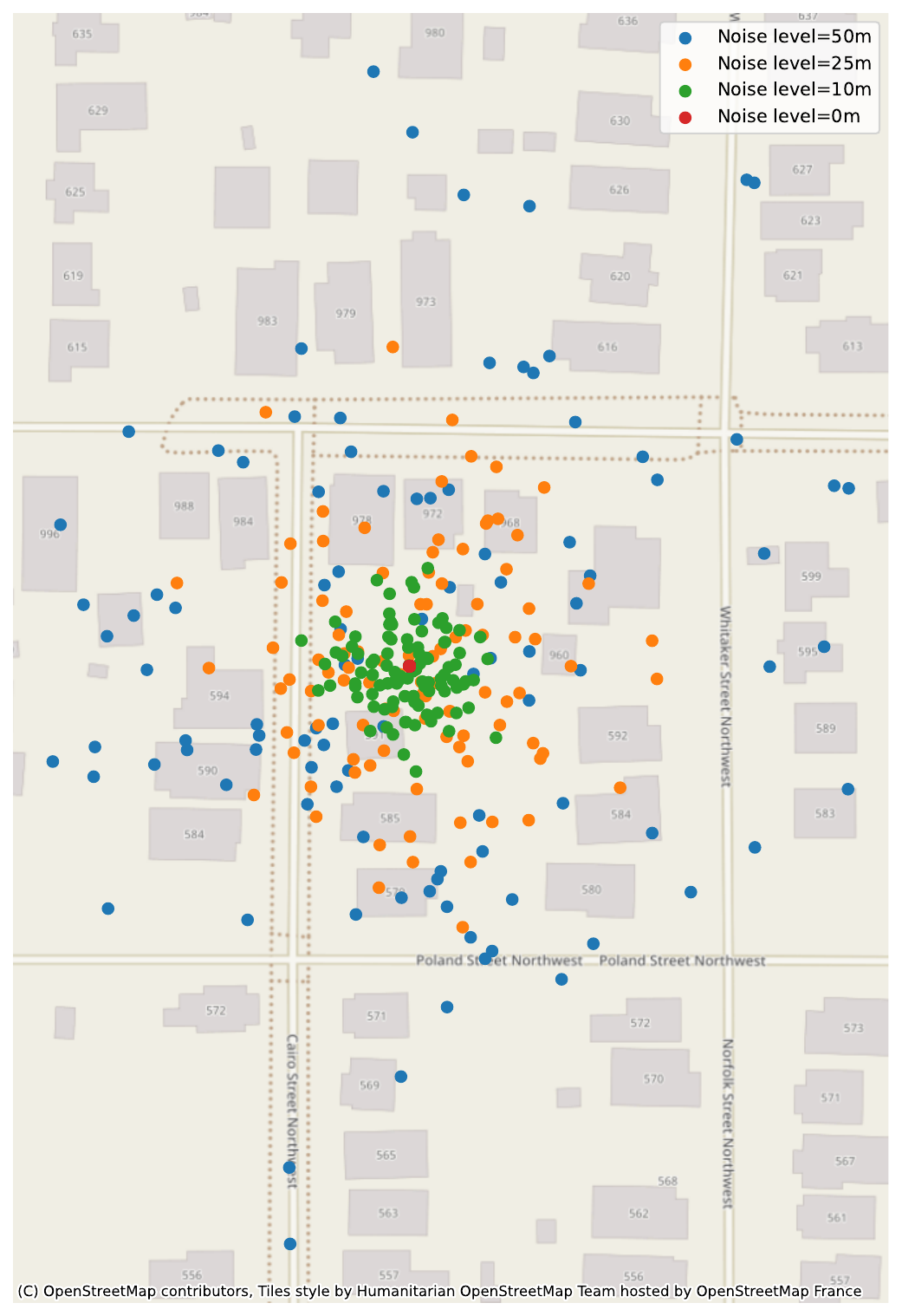}
    \caption{Observed locations for Agent 0 during a staypoint under different noise levels}
    \label{fig:staypoint}
\end{figure}

\begin{figure}
    \centering
    \includegraphics[width=0.9\linewidth,height=6cm]{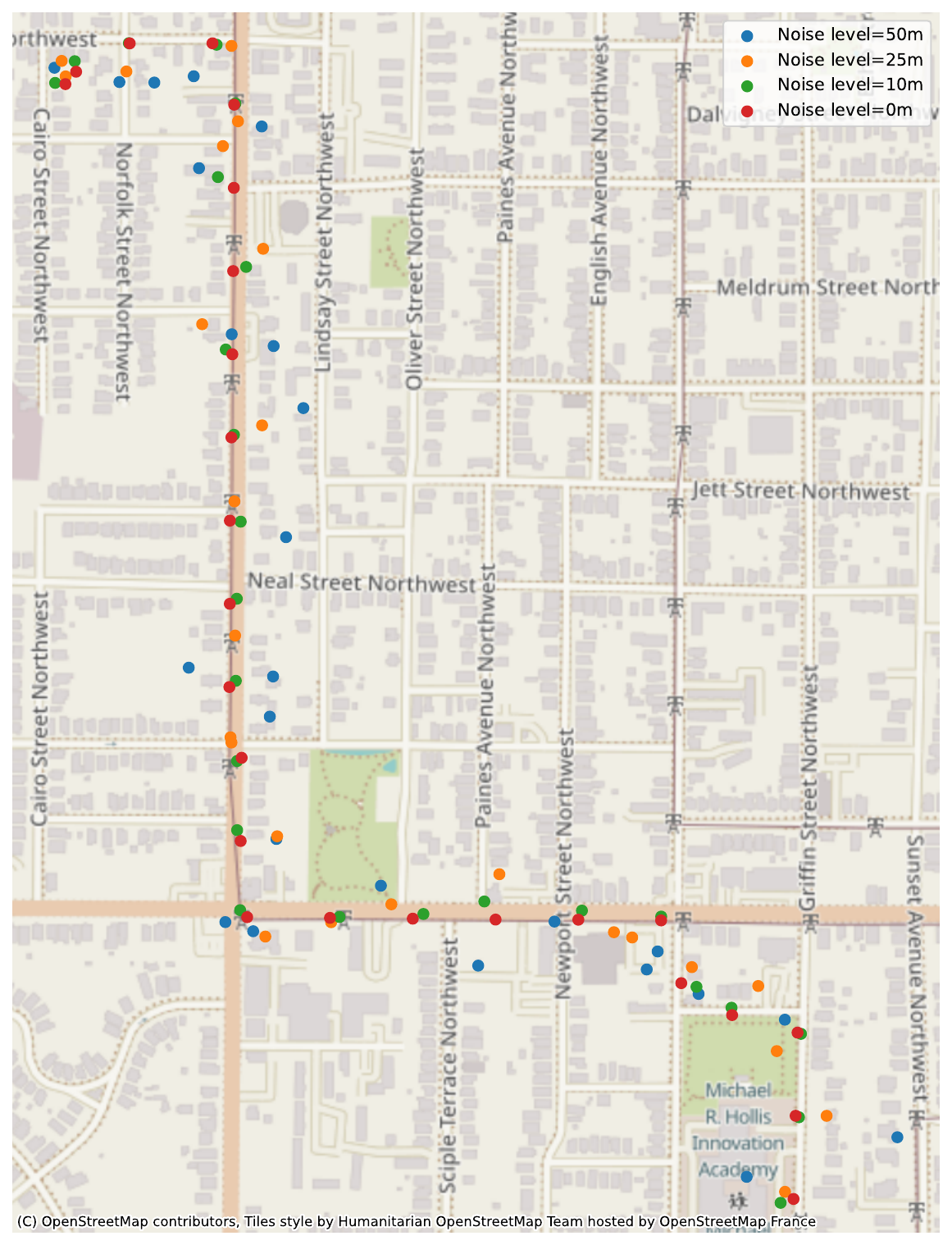}
    \caption{Observed locations for Agent 0 during travel under different noise levels}
    \label{fig:travel}
\end{figure}

The simulated trajectories have no noise, so while an agent is at a staypoint, every observation will have identical latitudes and longitudes. To observe how staypoint detection methods perform under realistic noise, we apply i.i.d. Gaussian noise to the locations at different intensities that is:
$$
\mbox{Noisify}[\mbox{lat},\mbox{long}]=[\mbox{lat}+N(0,\frac{\mbox{dist}^2}{0.5\pi}),\mbox{long}+N(0,\frac{\mbox{dist}^2}{0.5\pi})],
$$ 
where $\mbox{Noisify}$ is the noise function, $(\mbox{lat},\mbox{long})$ are the un-noised coordinates, $N(0,\sigma^2)$ is a normal-distributed random variable having a mean of $0$ with a variance of $\frac{\mbox{dist}^2}{0.5\pi}$, where $\mbox{dist}$ is a parameter that denotes the expected distance of the generated point to the un-noised point. For our experiments, we apply three difference levels of noise having $\mbox{dist}\in\{10m,25m,50m\}$. This noisification is applied independently to each data point in the dataset. Including the un-noised data, this gives us four datasets with different intensities of noise. We can observe these noise levels closely in Figures~\ref{fig:staypoint} and~\ref{fig:travel}, which show a portion of the trajectory for Agent 0. Figure~\ref{fig:staypoint} shows observed locations during a staypoint, where there is no variation without noise, and as the noise levels increase, so does the average distance of the observed points to the true location. Figure~\ref{fig:travel} shows the same thing but during a travelling period, where the agents locations are found directly on a road without noise, with the travelled path becoming harder to determine under increasing noise levels. Finally, we observe the full trajectory of agent 0 under different noise levels in Figure~\ref{fig:traj}. The density of points allows us to see several staypoints, as well as paths which are travelled more or less often than others. We also observe a staypoint in the southwest corner which has no path in or out of it - this a simulation artefact occuring in cases where the exists no path to the destination, often caused by discontinuities of the OpenStreetMap road and walkway networks. 

\begin{figure*}[t]
    \centering

    \begin{minipage}{0.32\textwidth}
        \centering
        \includegraphics[width=\linewidth]{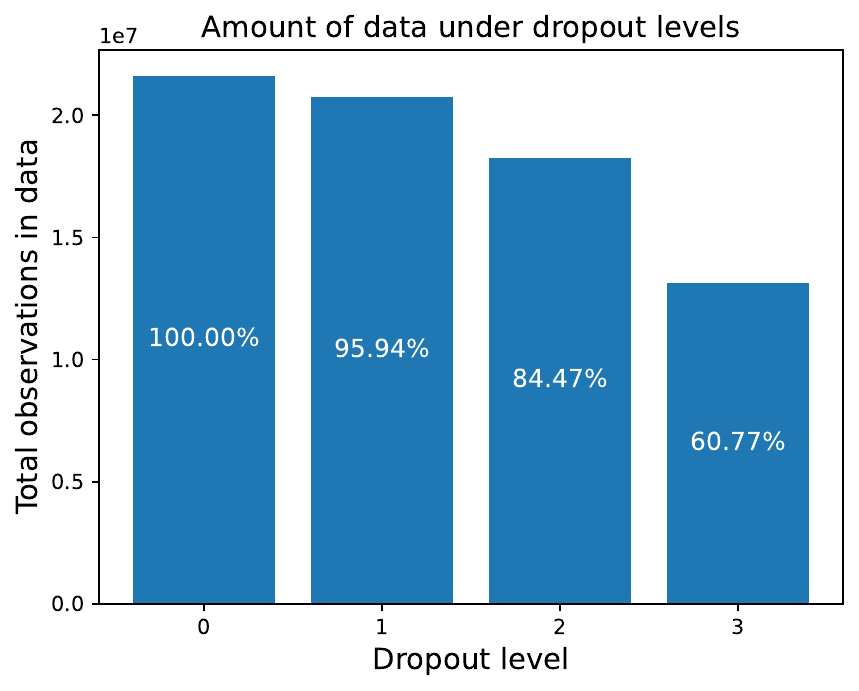}
        \caption*{(a) Number of total dataset rows under different dropout levels}
        \label{fig:dropout}
    \end{minipage}
    \hfill
    \begin{minipage}{0.32\textwidth}
        \centering
        \includegraphics[width=\linewidth]{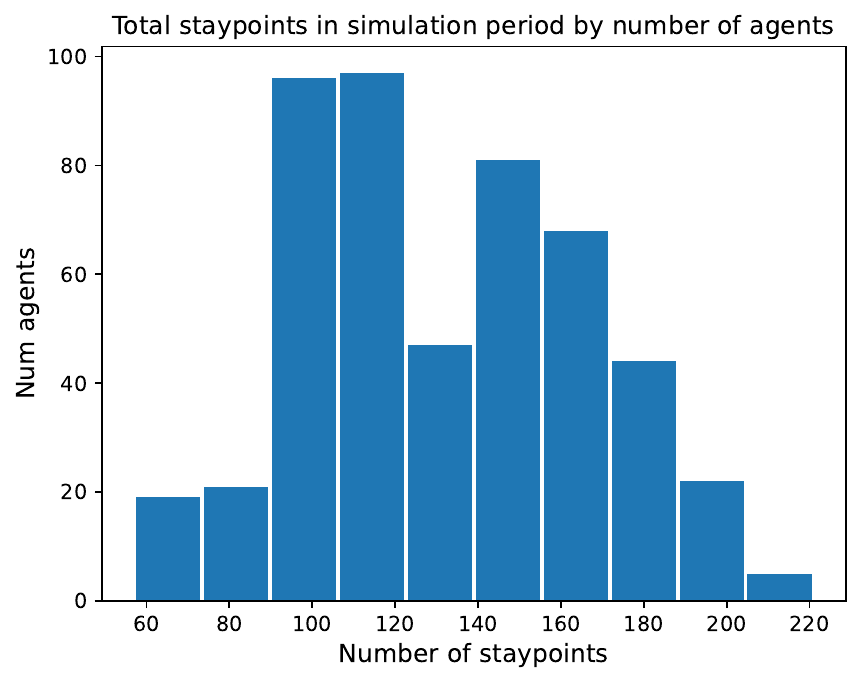}
        \caption*{(b) Binned number of staypoints for each agent}
        \label{fig:sps_hist}
    \end{minipage}
    \hfill
    \begin{minipage}{0.32\textwidth}
        \centering
        \includegraphics[width=\linewidth]{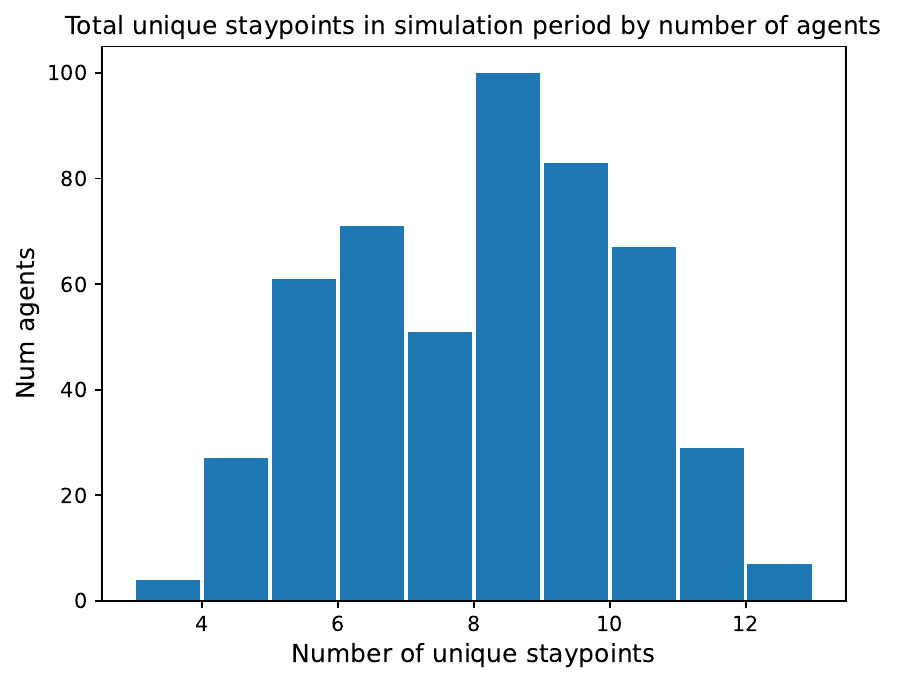}
        \caption*{(c) Binned number of unique staypoints for each agent}
        \label{fig:unique_sps_hist}
    \end{minipage}

    \caption{Dataset size and staypoint distributions under different dropout and staypoint aggregation settings.}
    \label{fig:dropout_staypoint_stats}
\end{figure*}

Further, we introduce data dropout to the trajectories. In real-world scenarios, missing data is one of the most common issues of trajectory datasets, due to both device failures and user error. By introducing known dropouts, we can assess how different methods handle this issue. When applying data dropout to a trajectory, we use two parameters- first, the expected number of dropouts per day (sampled from a Poisson distribution), and expected dropout duration (sampled from an exponential distribution). We apply three different dropout levels, with parameters as follows: \{(1, 1), (2, 2), (3, 4)\}. In the first case, we expect an agent to have one dropout per day with a duration of one hour, and in the last case, we expect three dropouts per day with a duration of four hours, so roughly half the trajectory is omitted from the data. Including the case with no data dropout, this gives us four levels of dropout intensity. Figure~\ref{fig:dropout_staypoint_stats} (a) demonstrates the amount of data in the dataset under these different dropout levels. We have over 2 million total observations in the full dataset with no dropout, and lose nearly 40\% of it under the harshest levels of dropout.

For each of the four noised datasets, we introduce data dropouts at each of the four levels of intensity, for a total of 16 trajectory datasets with a unique degree of noise and data dropout intensity, each of which are used as input to various staypoint detection methods. These datasets are simulated with realistic behavior and have realistic noise applied to them, allowing us to assess the quality of different staypoint detection methods against ground truth for the first time. In future work, these datasets may also vary the sampling rate, to assess how this affects the quality of the detected staypoints.



Figure~\ref{fig:dropout_staypoint_stats} (b) shows a histogram binning the number of total observed staypoints per agent in the simulation by the number of agents in that range. We see that over the 30-day period, it is most common for an agent to visit between 90 and 120 staypoints- or roughly three to four locations in an average day. But, less active agents may visit as few as 60 staypoints, and very active agents may visit as many as 200- or over six locations a day, on average. Then in Figure~\ref{fig:dropout_staypoint_stats} (c), we see a similar histogram, but counting the number of unique staypoints visited per agent. We observe that a typical agent visits around nine unique staypoint locations during the simulation period. These will typically include one home location, one work location, and several unique restaurant or recreation locations. Finally, we observe that on average, the 500 agents spend 29 days and five hours at stationary locations, and only about 19 total hours in transit.

\section{Methodologies}
\label{sec:methodology}
We provide a brief description of each method used in this paper. Full implementation details can be found on our \href{https://github.com/amirih/staypoint}{Github} page: https://github.com/amirih/staypoint.

Staypoint detection is typically an unsupervised problem, and as such we provide several unsupervised methods, discussed in Section~\ref{sec:single-shot}. In these cases, a universal parameter setting is used across the 16 datasets, which we call ``single-shot" approaches. On the other hand, since we have ground-truth, we may also fine-tune the parameters of any given method to achieve the best ideal results for that method on a given dataset. We do so with some methods, here, discussed in Section~\ref{sec:param-optimized}. Since the methods themselves are no supervised, but still use the ground-truth to fine-tune parameters, we call these methods ``parameter-optimized." When possible, we try to provide both a single-shot and parameter-optimized result for any given method. Finally, we present a single fully-supervised method, possible in simulated datasets such as these, discussed in Section~\ref{sec:supervised}. 

\subsection{Single-shot Approaches} \label{sec:single-shot}

\subsubsection{Hybrid Stay Window (HSW)}

The Hybrid Stay Window (HSW) method is designed to detect stay points from individual trajectory records by combining an anchored window-expansion strategy with spatial and temporal stay-point constraints. HSW extends the stay-point detection algorithm proposed in~\cite{li2008mining} by refining the candidate window boundary, reducing redundant trajectory scans, and introducing noise tolerance. Specifically, the method expands a candidate stay window from a fixed starting observation and determines whether the accumulated observations represent a meaningful stationary episode. To improve robustness, HSW can ignore a limited number of observations that temporarily fall outside the spatial threshold before terminating the candidate window.

The method first preprocesses the trajectory records by removing incomplete observations, converting timestamps into a consistent temporal format, and ordering each individual's trajectory chronologically. After preprocessing, the algorithm uses a spatial threshold to define the movement radius within which an individual is considered stationary and a temporal threshold to define the minimum duration required for a valid stay point.

HSW processes each individual trajectory independently. For each trajectory, the algorithm begins at a starting observation and gradually expands a candidate window forward in time. The window continues to grow while later observations remain within the spatial threshold of the starting location. If an observation falls outside this radius, the algorithm may ignore it as noise, up to a predefined noise tolerance limit. When the number of outside-radius observations exceeds this limit, the algorithm evaluates the accumulated window up to the previous observation. If the duration of this candidate window is at least the temporal threshold, the window is classified as a stay point. The stay-point location is represented by the average latitude and longitude of all observations in the window, while the arrival time, departure time, duration, number of points, and spatial radius are recorded. After detecting a valid stay point, the algorithm advances to the first observation outside the detected window, reducing redundant checks and improving efficiency. If the candidate window does not satisfy the temporal threshold, the algorithm advances the starting point by one observation.

This approach extends a standard stay-point sliding-window baseline in two main ways. First, it applies a boundary-aware jump strategy after detecting a stay point, ensuring that observations already assigned to a valid stay region are not repeatedly scanned. Second, it introduces noise tolerance, allowing the candidate window to remain valid despite a small number of temporary deviations outside the spatial threshold. Further details are provided in Appendix~\ref{app:algorithm_details}, Algorithm~\ref{alg:hsw}. 
\subsubsection{Temporal DBSCAN}
The ST-DBSCAN algorithm~\cite{BIRANT2007208} clusters spatial-temporal data together to identify staypoints, utilizing the original DBSCAN algorithm~\cite{ester1996density}. We propose a similar temporal DBSCAN (T-DBSCAN) which utilizes DBSCAN on three dimensions: latitude, longitude, and time. In this algorithm, time is transformed into a numerical value as the number of seconds since the onset of the trajectory, and all three dimensions are normalized to the range 0-1.  This algorithm requires two parameters, $minPts$ and $\epsilon$, which function similarly to the original DBSCAN algorithm, and a minimum staypoint time $minTime$. We select $minPts$ as a function of $minTime$ and the sampling rate of 1 minute per observation. Notably, selecting $minPts$ this way is considerably harder when the sampling rate is variable, as is common in real-world data. Future work may vary the $minPts$ depending on the local sampling rate.

After clusters are detected they may not be temporally contiguous, and they may fail to meet our $minTime$ threshold. We employ some post-processing steps to resolve these issues, detailed in in Appendix~\ref{app:algorithm_details}, Algorithm~\ref{alg:tdbscan}. Essentially, for each detected cluster, we observe its temporal gaps. If that gap has a duration of at least $minTime$, then we split the cluster into two, and otherwise we simply add the gap to the cluster. Finally, we check that our resulting clusters have a duration of at least $minTime$, and remove them otherwise. For the resulting staypoint clusters, we simply take the mean location. This post-processing is non-deterministic. That is, if two detected staypoints overlap temporally, then the result will depend on the order in which they are processed.

We provide a single-shot version of Temporal DBSCAN where consistent parameters are used across all experiments, which were tuned to a particular dataset through human efforts. Then, using the ground truth in our experiments, we use grid search to also provide parameter-optimized results. 

\subsubsection{Sequential Stay Point Extraction (SSPE)}
To reliably isolate stationary episodes from raw trajectory streams afflicted by extreme stochastic spatial degradation and protracted temporal occlusions, we propose the Sequential Stay Point Extraction Algorithm (SSPE).
Traditional stay point detection methods are fundamentally ill-suited for highly sparse or noisy data. Their reliance on global spatial density frequently causes them to erroneously fracture continuous stationary periods in the presence of large temporal gaps or high spatial deviation.
In contrast, SSPE employs a chronological, state-driven approach that processes trajectories sequentially to preserve spatiotemporal continuity.
Specifically, the algorithm maintains a dynamic spatial centroid and evaluates incoming coordinates against defined spatiotemporal thresholds.
To mitigate spatial anomalies caused by noise, SSPE utilizes an $n$-step lookahead heuristic that tolerates transient coordinate drift, provided the trajectory rapidly reverts to the previously established centroid. Concurrently, a temporal bridging mechanism spans missing data intervals up to a maximum dropout threshold ($\tau_{max}$), ensuring that long stationary periods are not artificially split by temporary signal dropouts. More details on this method are provided in Appendix~\ref{app:algorithm_details}, Algorithm~\ref{alg:sspe}.

\subsubsection{Hidden Markov Model with Global Emission Learning}

We adopt a two-state Hidden Markov Model (HMM) to segment GPS trajectories into \emph{stop} and \emph{move} behavioral states, following a widely used formulation for movement-state inference in sequential data \cite{rabiner2002tutorial,patterson2009classifying}. Given a time-ordered trajectory, each time step is represented by a $K$-dimensional observation vector composed of multi-step spatial displacements, where the $k$-th component corresponds to the distance between locations separated by $k$ time indices. Feature components requiring unavailable future positions are treated as missing and excluded from likelihood evaluation. This multi-step representation provides a temporally smoothed description of movement while remaining compatible with standard probabilistic sequence models.

State-dependent emission probabilities are modeled using diagonal multivariate Gaussian distributions over the $K$-step displacement features. Rather than fixing emission parameters heuristically, we estimate them using a global Expectation--Maximization (EM) procedure \cite{dempster1977maximum} applied to the concatenation of all trajectories, while holding the initial state distribution and transition probabilities fixed. In the E-step, posterior state responsibilities are computed using the forward--backward algorithm for hidden Markov models \cite{baum1970maximization}, and in the M-step, emission means and variances are updated via responsibility-weighted sufficient statistics. After global parameter learning, each trajectory is decoded independently using the Viterbi algorithm to obtain the most likely state sequence \cite{forney2005viterbi}. Staypoints are then defined as contiguous segments labeled as the stop state whose duration exceeds a predefined minimum dwell-time threshold, consistent with established definitions in trajectory mining and location analysis \cite{zheng2009mining}.

\subsubsection{Centroid-Based Sliding Window}

We include a centroid-based variant of the classical sliding-window staypoint detector. Unlike the standard fixed-anchor formulation, which compares each new point to the first point of the candidate window, this method maintains the running centroid of the current candidate stay region and compares incoming points to that centroid. If the incoming point remains within the spatial threshold, it is added to the candidate window; otherwise, the method evaluates whether the accumulated window satisfies the minimum dwell-time threshold.

For each detected staypoint, the method reports the mean latitude and longitude of the accepted window, the first and last timestamps as arrival and departure times, the duration, and the number of trajectory points. This method is unsupervised and uses fixed spatial and temporal thresholds, but it differs from the Trackintel-style fixed-anchor baseline by allowing the reference location of the candidate stay region to move with the local centroid.

\subsubsection{Trackintel Baseline} \label{sec:ti_baseline}
As an single-shot baseline, we implement the Sliding Window Staypoint Detection Algorithm~\cite{li2008mining}, one of the most widely used staypoint detection algorithm, via the Trackintel library in Python. We apply realistic estimated parameters which were tuned by human supervision to a similar dataset with no ground-truth.

\subsubsection{Trackintel with Dropout Filter}
The Sliding Window Staypoint Detection Algorithm has a known issue with data dropout which results in spurious staypoints being generated, and~\cite{kennedy2025improvement} provides a solution to this by filtering out predicted staypoints which coincide with data dropout. We implement this using Trackintel with the same parameters as in Section~\ref{sec:ti_baseline}, but add the dropout filter to test its improvement.

\subsubsection{Adaptive-Radius Sliding Window (ASW)}

We implement an adaptive-radius sliding window algorithm inspired by the classical sliding-window approach used in staypoint detection methods, while dynamically adjusting the spatial threshold according to the local spatial variability of the trajectory. Unlike fixed-distance methods, our approach adapts the allowable stay radius based on the local movement noise around each trajectory point, making it more robust to heterogeneous movement behaviors and GPS uncertainty.

Given a trajectory sorted temporally, the algorithm first projects latitude and longitude coordinates into a local Cartesian coordinate system measured in meters. For each trajectory point \(i\), a local spatial variability measure is estimated using a rolling window of neighboring points. Specifically, we compute the standard deviation of the projected coordinates within a local window of size \(w\), and define the local spatial noise as
\[
\sigma_i = \sqrt{\sigma_x(i)^2 + \sigma_y(i)^2},
\]
where \(\sigma_x(i)\) and \(\sigma_y(i)\) denote the standard deviation of the projected \(x\)- and \(y\)-coordinates around point \(i\), respectively.

Using this local variability estimate, the adaptive spatial threshold for point \(i\) is defined as
\[
r_i = \max(r_{\min}, \alpha \cdot \sigma_i),
\]
where \(r_{\min}\) is a minimum allowed radius and \(\alpha\) is a scaling factor controlling the sensitivity of the adaptive radius to local trajectory noise.

The staypoint detection itself follows a two-pointer scanning strategy. Starting from an anchor point \(i\), the algorithm incrementally expands a window by moving pointer \(j\) forward in time while the haversine distance between points \(i\) and \(j\) remains within the adaptive radius \(r_i\). Once the distance exceeds \(r_i\), the temporal duration of the candidate segment is evaluated. If the duration exceeds a predefined temporal threshold \(t_{\min}\), the segment is identified as a staypoint.

For each detected staypoint, the centroid coordinates are computed as the mean latitude and longitude of all points within the segment. The algorithm additionally records the arrival time, leave time, duration, number of trajectory points, and the adaptive radius used for detection.

Compared to fixed-radius staypoint detection approaches, this adaptive formulation allows the method to automatically enlarge the spatial threshold in noisy or highly dynamic regions while maintaining tighter thresholds in stable regions. As a result, the algorithm is better suited for trajectories exhibiting heterogeneous sampling quality and varying mobility behaviors.

\subsection{Parameter-Optimized Approaches} \label{sec:param-optimized}

\subsubsection{Trackintel Hyperband Search}

Trackintel~\cite{martin2023trackintel} is a common method used in staypoint detection. Beyond the single-shot Trackintel baseline, we applied Hyperband~\cite{li2018hyperband}, an efficient parameter optimization algorithm, to search for locally optimized parameter settings for our datasets. Three parameters used by Trackintel were searched in this process: \textit{time}, \textit{distance}, and \textit{gap}. Here, \textit{time} represents the minimum amount of time a person must remain within a small area before Trackintel identifies it as a staypoint; \textit{distance} represents the maximum movement range that can still be treated as belonging to the same staypoint; and \textit{gap} represents the largest allowed time interval between consecutive GPS points when constructing staypoints, where a gap above this threshold is treated as missing data rather than part of the same staypoint.

We implemented the parameter search as a three-stage process with different subsets of data used in each stage: stage 1 uses 25\%, stage 2 uses 75\%, and stage 3 uses 100\%. With this design, the parameters can be quickly searched over a large range of combinations. The search ranges we defined are \textit{distance} (\textit{dist}): 50--200 meters, \textit{time} (\textit{t}): 5--20 minutes, and \textit{gap}: 30--180 minutes.

After several exploratory runs, we settled with the following setting in an attempt to balance exploration and the discovery of good parameter combinations. In stage 1, we initially randomly selected 36 candidate combinations, evaluated their performance based on F1 score, and ranked them. We kept the top 12 candidates, and then used the parameter ranges covered by these top candidates to redefine the search ranges of the three parameters. Based on the new ranges, we reselected 10 candidate combinations to explore more possible combinations, keeping the top 2 candidates from the previous iteration for a total of 12, and ranked them again. This process was repeated for 10 iterations, and the last set of 12 candidates were sent into the next stage. In stage 2, a similar process to stage 1 was performed, but with a larger 75\% dataset. We selected the top 4 candidates for iteration, kept the best single candidate, and reselected three new candidate combinations within the newly defined search range, with the last set of 4 candidates sent into the next stage. This process was repeated for 10 iterations. In the last stage, all 4 candidates were evaluated on the full dataset, and 5 additional iterations were performed to further optimize their performance on all data. In this way, we searched the parameter space we defined and treated the final result as a locally optimal setting.

\vspace{-0.4cm}
\subsection{Supervised Approaches} \label{sec:supervised}

\begin{table*}[t]
\centering
\caption{Spatio-Temporal Similarity results across the PoL dataset with different noise and dropout levels from various methods}
\label{tab:f1-results}
\small
\setlength{\tabcolsep}{3pt}
\resizebox{\textwidth}{!}{
\begin{tabular}{c cccc cccc cccc cccc}
\toprule

Noise level
& \multicolumn{4}{c}{0m} 
& \multicolumn{4}{c}{10m} 
& \multicolumn{4}{c}{25m} 
& \multicolumn{4}{c}{50m} \\

\cmidrule(lr){2-5} \cmidrule(lr){6-9} \cmidrule(lr){10-13} \cmidrule(lr){14-17}

Dropout Level
& 0 & 1 & 2 & 3
& 0 & 1 & 2 & 3
& 0 & 1 & 2 & 3
& 0 & 1 & 2 & 3 \\

\midrule

\multicolumn{17}{c}{\textbf{Single-Shot (Unsupervised) Approaches}} \\

\midrule

T-DBSCAN
& 0.959828 & 0.902150 & 0.732224 & 0.479151 
& 0.954635 & 0.896001 & 0.722184 & 0.461099 
& 0.918988 & 0.856432 & 0.691742 & 0.435220 
& \textbf{0.744952} & \textbf{0.673614} & \textbf{0.511763} & \textbf{0.296200} \\

SSPE
& \textbf{0.9943} & \textbf{0.9403} & \textbf{0.7990} & \underline{\textbf{0.5632}} 
& \underline{\textbf{0.9902}} & \underline{\textbf{0.9348}} & \underline{\textbf{0.7938}} & \underline{\textbf{0.5605}} 
& \underline{\textbf{0.9835}} & \underline{\textbf{0.9280}} & \underline{\textbf{0.7935}} & \underline{\textbf{0.5563}} 
& 0.9044 & 0.8553 & 0.7299 & 0.5171 \\ 

3-Step HMM-GEM
& \underline{\textbf{0.996878}} & \underline{\textbf{0.940353}} & \underline{\textbf{0.800760}} & \textbf{0.561466} 
& \textbf{0.971279} & \textbf{0.915559} & \textbf{0.781775} & \textbf{0.549846} 
& \textbf{0.936862} & \textbf{0.883671} & \textbf{0.762270} & \textbf{0.535331} 
& \underline{\textbf{0.834935}} & \underline{\textbf{0.790557}} & \underline{\textbf{0.683096}} & \underline{\textbf{0.487092}} \\

Hybrid Stay Window
 & 0.9775  & 0.925  & 0.791  & 0.5565 
 & 0.9712  & 0.918  & 0.7844  & 0.5523 
 & 0.7415  & 0.7063  & 0.618  & 0.4508 
 & 0.0262  & 0.0259  & 0.0253  & 0.0241  \\

Trackintel Baseline
& 0.950141 & 0.813623 & 0.654156 & 0.472968
& 0.946860 & 0.808218 & 0.648805 & 0.466290
& 0.019170 & 0.018944 & 0.018709 & 0.017666
& 0.000481 & 0.000493 & 0.000523 & 0.000433 \\

Trackintel w/Filter
& 0.950141 & 0.813884 & 0.654508 & 0.473100
& 0.946860 & 0.808485 & 0.648914 & 0.466331
& 0.019170 & 0.018952 & 0.018705 & 0.017651
& 0.000481 & 0.000448 & 0.000459 & 0.000401 \\

  Centroid Sliding Window
  & 0.9429 & 0.8856 & 0.7439 & 0.5183
  & 0.9431 & 0.8844 & 0.7496 & 0.5247
  & 0.6347 & 0.6037 & 0.5317 & 0.3828
  & 0.0043 & 0.0043 & 0.0043 & 0.0041 \\
  
\midrule

\multicolumn{17}{c}{\textbf{Parameter-Optimized Approaches}} \\

\midrule

T-DBSCAN (Grid Search)
& 0.969726 & 0.904775 & 0.752909 & 0.498476 
& 0.955766 & 0.898982 & 0.741283 & 0.486296 
& 0.927948 & 0.873436 & 0.729121 & 0.471568 
& \textbf{0.844966} & \textbf{0.788836} & \textbf{0.643466} & 0.389943 \\

SSPE (Grid search)
& \underline{\textbf{0.9952}} & \underline{\textbf{0.9407}} & \underline{\textbf{0.8029}} & \underline{\textbf{0.5632}} 
& \underline{\textbf{0.9905}} & \underline{\textbf{0.9348}} & \underline{\textbf{0.7938}} & \underline{\textbf{0.5605}} 
& \underline{\textbf{0.9835}} & \underline{\textbf{0.9280}} & \underline{\textbf{0.7935}} & \underline{\textbf{0.5563}} 
& \underline{\textbf{0.9302}} & \underline{\textbf{0.8783}} & \underline{\textbf{0.7476}} & \underline{\textbf{0.5256}} \\

Trackintel Hyperband Search
& \textbf{0.9875} & 0.9257 & 0.7709 & \underline{\textbf{0.5632}}
& \textbf{0.9799} & 0.9163 & 0.7639 & 0.5352
& \textbf{0.9497} & \textbf{0.8892} & \textbf{0.7477} & \textbf{0.5240}
& 0.6664 & 0.6310 & 0.5322 & 0.3984 \\

Hybrid Stay Window
 & 0.987  & \textbf{0.9335} & \textbf{0.7975}  & \textbf{0.5603} 
 & 0.9712  & \textbf{0.9182}  & \textbf{ 0.7848 } & \textbf{0.5531} 
 & 0.8991  & 0.8501  & 0.7335  & 0.5139 
 & 0.7255  & 0.6898  & 0.5993  & \textbf{0.4362} \\

Adaptive-radius Sliding Window
& 0.9836 & 0.9301 & 0.7949 & 0.5581
& 0.9677 & 0.9140 & 0.7813 & 0.5513 
& 0.6884 & 0.6473 & 0.5589 & 0.3900 
& 0.6153 & 0.5836 & 0.5042 & 0.3653 \\

\midrule

\multicolumn{17}{c}{\textbf{Fully Supervised Approaches}} \\

\midrule

Histogram Gradient Boosting
  & 0.9981 & 0.9467 & 0.8142 & 0.5921
  & 0.9713 & 0.9291 & 0.7804 & 0.5623
  & 0.9826 & 0.9218 & 0.7877 & 0.4702
  & 0.9520 & 0.8914 & 0.7706 & 0.5313 \\
\bottomrule
\end{tabular}
}

\end{table*}

\begin{table*}
\centering
\caption{Temporal IoU score results across the PoL dataset with different noise and dropout levels from various methods }
\label{tab:temporal-iou-results}
\small
\setlength{\tabcolsep}{3pt}
\resizebox{\textwidth}{!}{
\begin{tabular}{c cccc cccc cccc cccc}
\toprule

Noise level
& \multicolumn{4}{c}{0m} 
& \multicolumn{4}{c}{10m} 
& \multicolumn{4}{c}{25m} 
& \multicolumn{4}{c}{50m} \\

\cmidrule(lr){2-5} \cmidrule(lr){6-9} \cmidrule(lr){10-13} \cmidrule(lr){14-17}

Dropout Level
& 0 & 1 & 2 & 3
& 0 & 1 & 2 & 3
& 0 & 1 & 2 & 3
& 0 & 1 & 2 & 3 \\

\midrule

\multicolumn{17}{c}{\textbf{Single-Shot (Unsupervised) Approaches}} \\

\midrule

T-DBSCAN
& 0.9723 & 0.9573 & 0.8785 & 0.7366 
& 0.9699 & 0.9542 & 0.8706 & 0.7061 
& \textbf{0.9445} & \textbf{0.9235} & 0.8398 & 0.6765 
& 0.8055 & 0.7721 & 0.6683 & 0.4819 \\

SSPE
& \underline{\textbf{0.9926}} & \underline{\textbf{0.9816}} & \underline{\textbf{0.9402}} & \underline{\textbf{0.8564}} 
& \underline{\textbf{0.9787}} & \underline{\textbf{0.9676}} & \underline{\textbf{0.9277}} & \underline{\textbf{0.8437}} 
& \underline{\textbf{0.9757}} & \underline{\textbf{0.9645}} & \underline{\textbf{0.9255}} & \underline{\textbf{0.8404}} 
& \underline{\textbf{0.9318}} & \underline{\textbf{0.9228}} & \underline{\textbf{0.8857}} & \underline{\textbf{0.8089}} \\

3-Step HMM-GEM
& 0.9448 & 0.9347 & 0.8996 & 0.8159 
& 0.9469 & 0.9367 & 0.9028 & 0.8185 
& 0.9280 & 0.9185 & \textbf{0.8878} & \textbf{0.8071} 
& \textbf{0.8472} & \textbf{0.8401} & \textbf{0.8148} & \textbf{0.7459} \\

Hybrid Stay Window
& \textbf{0.9769} & \textbf{0.9671} & \textbf{0.9325} & \textbf{0.8488} 
& \textbf{0.9710} & \textbf{0.9612} & \textbf{0.9274} & \textbf{0.8430} 
& 0.7506 & 0.7467 & 0.7286 & 0.6767 
& 0.0626 & 0.0639 & 0.0664 & 0.0703 \\

Trackintel Baseline
& 0.9547 & 0.8897 & 0.8162 & 0.7558 
& 0.9535 & 0.8863 & 0.8129 & 0.7499 
& 0.0540 & 0.0539 & 0.0537 & 0.0530 
& 0.0180 & 0.0184 & 0.0184 & 0.0184 \\

Trackintel w/Filter
& 0.9547 & 0.8907 & 0.8176 & 0.7568 
& 0.9535 & 0.8874 & 0.8137 & 0.7506 
& 0.0540 & 0.0539 & 0.0536 & 0.0530 
& 0.0180 & 0.0181 & 0.0181 & 0.0181 \\

Centroid Sliding Window
& 0.9460 & 0.9324 & 0.8961 & 0.8188 
& 0.9460 & 0.9322 & 0.8962 & 0.8173 
& 0.7290 & 0.7238 & 0.7093 & 0.6670 
& 0.0288 & 0.0302 & 0.0334 & 0.0394 \\
\midrule

\multicolumn{17}{c}{\textbf{Parameter-Optimized Approaches}} \\

\midrule

T-DBSCAN (Grid Search)
& 0.9779 & 0.9524 & 0.9013 & 0.7686 
& 0.9648 & 0.9573 & 0.8869 & 0.7498 
& \textbf{0.9589} & \textbf{0.9484} & 0.8905 & 0.7432 
& \textbf{0.9134} & \textbf{0.8968} & \textbf{0.8318} & 0.6306 \\

SSPE  (Grid Search)
& \underline{\textbf{0.9910}} & \underline{\textbf{0.9811}} & \underline{\textbf{0.9429}} & \underline{\textbf{0.8564}} 
& \underline{\textbf{0.9777}} & \underline{\textbf{0.9676}} & \textbf{0.9278} & \textbf{0.8438} 
& \underline{\textbf{0.9758}} & \underline{\textbf{0.9645}} & \underline{\textbf{0.9255}} & \underline{\textbf{0.8405}} 
& \underline{\textbf{0.9511}} & \underline{\textbf{0.9410}} & \underline{\textbf{0.9031}} & \underline{\textbf{0.8222}} \\

Trackintel Hyperband Search
& 0.9608 & 0.9446 & 0.8911 & 0.8253 
& 0.9561 & 0.9445 & 0.8956 & 0.8155 
& 0.9484 & 0.9322 & 0.8877 & 0.8012 
& 0.6829 & 0.6751 & 0.6228 & 0.5899 \\

Hybrid Stay Window
& \textbf{0.9788} & \textbf{0.9680} & \textbf{0.9312} & \textbf{0.8456} 
& \textbf{0.9710} & \textbf{0.9634} & \underline{\textbf{0.9307}} & \underline{\textbf{0.8480}} 
& 0.9450 & 0.9363 & \textbf{0.9071} & \textbf{0.8264} 
& 0.7841 & 0.7792 & 0.7587 & \textbf{0.6847} \\

Adaptive-radius Sliding Window
& 0.9781 & 0.9673 & 0.9305 & 0.8449 
& 0.9621 & 0.9513 & 0.9174 & 0.8337 
& 0.8062 & 0.7973 & 0.7721 & 0.7106 
& 0.6913 & 0.6871 & 0.6756 & 0.6314 \\

\midrule

\multicolumn{17}{c}{\textbf{Fully Supervised Approaches}} \\

\midrule

Histogram Gradient Boosting
& 1.0 & 0.9918 & 0.9646 & 0.9100 
& 0.9865 & 0.9841 & 0.9393 & 0.8691 
& 0.9946 & 0.9795 & 0.9445 & 0.7737 
& 0.9807 & 0.9682 & 0.9409 & 0.8540 \\
\bottomrule
\end{tabular}
}

\end{table*}

\begin{table*}
\centering
\caption{Spatial overlap score results across the PoL dataset with different noise and dropout levels from various methods}
\label{tab:spatial-results}
\small
\setlength{\tabcolsep}{3pt}
\resizebox{\textwidth}{!}{
\begin{tabular}{c cccc cccc cccc cccc}
\toprule

Noise level
& \multicolumn{4}{c}{0m} 
& \multicolumn{4}{c}{10m} 
& \multicolumn{4}{c}{25m} 
& \multicolumn{4}{c}{50m} \\

\cmidrule(lr){2-5} \cmidrule(lr){6-9} \cmidrule(lr){10-13} \cmidrule(lr){14-17}

Dropout Level
& 0 & 1 & 2 & 3
& 0 & 1 & 2 & 3
& 0 & 1 & 2 & 3
& 0 & 1 & 2 & 3 \\

\midrule

\multicolumn{17}{c}{\textbf{Single-Shot (Unsupervised) Approaches}} \\

\midrule

T-DBSCAN
& 1.17e-05 & 1.17e-05 & 1.15e-05 & 1.09e-05 
& 2.10e-05 & 2.12e-05 & 2.20e-05 & 2.05e-05 
& 3.98e-05 & 3.99e-05 & 4.00e-05 & 4.10e-05 
& 8.91e-05 & 8.93e-05 & 9.16e-05 & 1.01e-04 \\

SSPE
& \textbf{4.30e-07} & \textbf{4.25e-07} & \textbf{1.27e-06} & \textbf{5.57e-07} 
& \textbf{1.15e-05} & \underline{\textbf{1.16e-05}} & 1.51e-05 & \textbf{1.25e-05}
& \underline{\textbf{2.84e-05}} & \underline{\textbf{2.85e-05}} & \underline{\textbf{2.88e-05}} & \textbf{3.03e-05} 
& \textbf{6.76e-05} & \textbf{6.79e-05} & \textbf{6.91e-05} & \textbf{6.93e-05} \\

3-Step HMM-GEM
& \underline{\textbf{3.86e-11}} & \underline{\textbf{4.14e-11}} & \underline{\textbf{3.80e-11}} & \underline{\textbf{4.18e-11}} 
& 1.16e-05 & \underline{\textbf{1.16e-05}} & \underline{\textbf{1.17e-05}} & \underline{\textbf{1.18e-05}} 
& \textbf{2.94e-05} & \textbf{2.95e-05} & \textbf{2.96e-05} & \underline{\textbf{2.98e-05}} 
& \underline{\textbf{6.62e-05}} & \underline{\textbf{6.65e-05}} & \underline{\textbf{6.69e-05}} & \underline{\textbf{6.77e-05}} \\

Hybrid Stay Window
& 4.30e-06 & 4.52e-06 & 4.24e-06 & 4.49e-06 
& 1.24e-05 & 1.25e-05 & 1.56e-05 & 1.26e-05 
& 3.22e-05 & 3.23e-05 & 3.25e-05 & 3.31e-05 
& 1.25e-04 & 1.25e-04 & 1.26e-04 & 1.27e-04 \\

Trackintel Baseline
& 5.87e-05 & 5.98e-05 & 6.24e-05 & 6.20e-05 
& \underline{\textbf{1.13e-05}} & 1.42e-05 & 1.53e-05 & 1.60e-05 
& 1.00e-04 & 1.01e-04 & 1.01e-04 & 1.01e-04 
& 2.28e-04 & 2.34e-04 & 2.35e-04 & 2.33e-04 \\

Trackintel w/Filter
& 5.87e-05 & 5.83e-05 & 5.82e-05 & 5.77e-05 
& \underline{\textbf{1.13e-05}} & \textbf{1.19e-05} & \textbf{1.25e-05} & 1.32e-05 
& 1.00e-04 & 1.00e-04 & 1.01e-04 & 1.01e-04 
& 2.28e-04 & 2.28e-04 & 2.29e-04 & 2.28e-04 \\

Centroid Sliding Window
& 3.01e-04 & 3.30e-04 & 3.95e-04 & 4.83e-04 
& 3.01e-04 & 3.33e-04 & 4.00e-04 & 4.86e-04 
& 2.67e-04 & 2.99e-04 & 3.46e-04 & 4.26e-04 
& 1.92e-04 & 2.05e-04 & 2.36e-04 & 2.98e-04 \\
\midrule

\multicolumn{17}{c}{\textbf{Parameter-Optimized Approaches}} \\

\midrule

T-DBSCAN (Grid Search)
& 9.61e-06 & 1.18e-05 & 1.19e-05 & 1.13e-05 
& 2.10e-05 & 2.14e-05 & 2.24e-05 & 2.07e-05 
& 3.93e-05 & 3.95e-05 & 3.93e-05 & \textbf{3.95e-05} 
& 7.51e-05 & 7.49e-05 & 7.48e-05 & 8.13e-05 \\

SSPE  (Grid Search)
& \underline{\textbf{4.45e-07}} & \underline{\textbf{4.36e-07}} & \underline{\textbf{1.28e-06}} & \underline{\textbf{5.57e-07}}
& \underline{\textbf{1.15e-05}} & \underline{\textbf{1.16e-05}} & \textbf{1.51e-05} & 1.25e-05 
& \underline{\textbf{2.84e-05}} & \underline{\textbf{2.85e-05}} & \underline{\textbf{2.88e-05}} & \underline{\textbf{3.03e-05}} 
& 6.74e-05 & \textbf{6.78e-05} & \textbf{6.89e-05} & \textbf{6.93e-05} \\

Trackintel Hyperband Search
& 6.19e-05 & 9.60e-05 & 1.32e-04 & 1.44e-04 
& \textbf{1.16e-05} & 4.69e-05 & 9.24e-05 & 8.64e-05 
& \textbf{2.89e-05} & 7.14e-05 & 1.02e-04 & 1.18e-04 
& \textbf{6.23e-05} & 8.75e-05 & 1.16e-04 & 1.21e-04 \\

Hybrid Stay Window
& \textbf{2.45e-06} & \textbf{2.56e-06} & \textbf{2.38e-06} & \textbf{2.25e-06} 
& 1.24e-05 & 1.23e-05 & 1.54e-05 & \textbf{1.23e-05} 
& 3.86e-05 & \textbf{3.84e-05} & \textbf{3.83e-05} & 3.98e-05 
& \underline{\textbf{6.21e-05}} & \underline{\textbf{6.25e-05}} & \underline{\textbf{6.29e-05}} & \underline{\textbf{6.51e-05}} \\

Adaptive-radius Sliding Window
& 2.64e-06 & 2.75e-06 & 2.58e-06 & 2.48e-06 
& \textbf{1.16e-05} & \textbf{1.17e-05} & \underline{\textbf{1.39e-05}} & \underline{\textbf{1.19e-05}} 
& 7.90e-05 & 8.13e-05 & 8.52e-05 & 9.30e-05 
& 1.14e-04 & 1.17e-04 & 1.21e-04 & 1.22e-04 \\

\midrule

\multicolumn{17}{c}{\textbf{Fully Supervised Approaches}} \\

\midrule

Histogram Gradient Boosting
& 1.30e-14 & 1.11e-04 & 4.41e-04 & 9.46e-04 
& 1.12e-05 & 1.12e-05 & 1.77e-04 & 4.18e-04 
& 2.80e-05 & 2.82e-05 & 2.49e-04 & 4.54e-04 
& 5.78e-05 & 6.35e-05 & 2.53e-04 & 5.43e-04 \\
\bottomrule
\end{tabular}
}

\end{table*}

\subsubsection{Histogram Gradient Boosting}

We include a supervised machine-learning baseline that formulates staypoint detection as point-level stop/move classification. For each trajectory point, we extract local motion and spatial-stability features, including previous and next displacement, rolling coordinate standard deviation, rolling mean and maximum displacement, distance to a local rolling centroid, and hour-of-day. Ground-truth staypoint intervals are used to label each trajectory point as either stop or move.

We train a histogram-based gradient boosting classifier using an agent-level split: the first 80\% of agents are used for training and the remaining 20\% are held out for testing. During training, all stop-labeled points are retained, while move-labeled points are downsampled to at most twice the number of stop points; classifier hyperparameters are fixed in code rather than tuned on the held-out agents. After point-level prediction, consecutive points classified as stop are merged into candidate staypoint segments, segments shorter than the minimum dwell-time threshold are discarded, and each remaining segment is converted into a staypoint using the mean latitude and longitude of its points and the first and last timestamps as arrival and departure times. Because this method uses ground-truth labels during training, we report it as a supervised baseline rather than as an unsupervised detector.

\vspace{-0.3cm}
\section{Experiments and Results}
\label{sec:results}
\subsection{Data}
For this experimental evaluation we use the dataset introduced in Section~\ref{sec:data} which captures 500 simulated users having 1-minute location updates over a 30-day period and having a ground truth of 65,873 staypoints. These simulated users have noise applied to the GPS signals at three different intensities for a total of four datasets including the un-noised version. These four datasets then each have data dropout applied to them at three different intensities, for a total of 16 datasets, including the versions with no data dropout. These different noise settings reflect real-world issues with trajectory data, and allow us to assess staypoint detection algorithms under different levels of noise.

\subsection{Metrics}
With little to no precedence for assessing detected staypoints against a ground truth, we have developed our own metrics for evaluating the quality of the detected staypoints.

\subsubsection{Spatio-Temporal Staypoint Similarity}\label{subsubsec:stsim}
Our first metric combines (1) the spatial distance between the detected and the ground truth staypoint and (2) the temporal distance between the detected start and end times of the staypoint against the ground truth staypoint. Intuitively, this metric defines the detected staypoints as a ``hit'' if it is both spatially and temporally close the the ground truth.

To formalize this intuition, a staypoint can be thought of as a 4-tuple $(t_{start}, t_{end}, x, y)$, consisting of the staypoints arrival and departure times, and its location in 2-dimensional space, most commonly as latitude and longitude. To determine whether a detected staypoint accurately reflects a ground truth staypoint, we simply check if each element of the tuple for the detected staypoint is within two thresholds, a spatial threshold $\tau_{s}$ serving as the maximum GPS distance between longitude or latitude to the ground truth, and a temporal threshold $\tau_{t}$ serving as the maximum minutes level difference between arrival or departure time to the ground truth. Then, with a staypoint $(Pred_{start}, Pred_{end}, Pred_{x}, Pred_{y})$ and its ground truth staypoint $(GT_{start}, GT_{end}, GT_{x}, GT_{y})$, we say the prediction is matched if and only if all the following conditions are correct:
\begin{align*}
    Pred_{x} &\in (GT_{x} - \tau_s, GT_{x} + \tau_s)\\
    Pred_{y} &\in (GT_{y} - \tau_s, GT_{y} + \tau_s)\\
    Pred_{start} &\in (GT_{start} - \tau_t, GT_{start} + \tau_t)\\
    Pred_{end} &\in (GT_{end} - \tau_t, GT_{end} + \tau_t)
\end{align*}
To be consistent, for the parameter-optimized approaches in this paper, we let the temporal thresholds to be 5 minutes, $\tau_t = 5$, and the spatial threshold to be 0.001 degrees (approximately 111 meters), $\tau_s = 0.001$.
%
%
Using this method, we compare detected staypoints against ground truth staypoints to quantify true/false hits/missed to get a recall score (results of which can be seen in Appendix~\ref{app:results} Table~\ref{tab:recall-results}), a precision score (results in Appendix~\ref{app:results} Table~\ref{tab:precision-results}, and F1 scores from these values, which we call Spatio-Temporal Similarity score for the purposes of this paper, using it as our primary metric which parameters were optimized for. We display results these from the 16 versions of the PoL dataset in Table~\ref{tab:f1-results}. 
With the same optimization parameters, we also evaluated the predictions with six more pairs of $\tau_s$ and $\tau_t$, showing the parameter sensitivity of the approaches by repeating our experiments with different levels of spatial and temporal accuracy. Due to page limitations, we present these results in our GitHub Repository.

This metric requires staypoint detection algorithms to account for data dropout effects. For instance, say a user spends 2 hours at a location, but after the first hour they cease transmitting signal for 15 minutes. An algorithm may identify this as a one-hour staypoint, then a 45-minute staypoint at the same location 15 minutes later. Both staypoints would be incorrect in this case, as the first one would have the wrong departure time and the second one would have the wrong arrival time. It is a difficult challenge in staypoint detection algorithms to account for this, and correctly merge the two staypoints into one. On the other hand, say that at the 2-hour staypoint, the user stops transmitting data after one hour, and never continues transmitting data. In this case, it is impossible to predict the departure time for the staypoint without additional trajectory prediction methods. 

\subsubsection{Temporal Similarity}\label{subsubsec:temporalsim}
Further, we propose a parameter-free metric to measure the temporal agreement of the detected staypoints with the ground truth staypoints. Intuitively, this metric measures the overlap of the time intervals during which a staypoint is detected but without consider the spatial correctness. For this purpose, we propose a Temporal Intersection over Union (IoU) score calculated as the length of temporal intersection of a predicted staypoint to its closest ground truth staypoint divided by their union, acting as a temporal-level Jaccard score. Formally, let $(Pred_{start},Pred_{end})$ denote the check-in and check-out time of a predicted staypoint, and $(GT_{start}, GT_{end})$ denote a ground truth staypoint having the best temporal overlap to this prediction, then equation for temportal IoU is as following:
\begin{align*}
    max(0, \frac{min(Pred_{end},GT_{end})-max(Pred_{start},GT_{start})}{max(Pred_{end},GT_{end}) - min(Pred_{start},GT_{start})})
\end{align*}
This metric is in the range [0, 1] with one being an exact matching and zero implying that detected and ground truth staypoint intervals are entirely disjoint. The average results for this metric are seen in Table~\ref{tab:temporal-iou-results}.


\subsubsection{Spatial Similarity}\label{subsubsec:spatialsim}
Finally, we propose a metric to measure the average distance between predicted and ground truth staypoints. We define spatial similarity as the distance between a predicted staypoint and the closest staypoint in the ground-truth having at least one second of temporal overlap. The average results are in Table~\ref{tab:spatial-results}, and were calulcated as Euclidean distance between (latitude, longitude) pairs where, again, a distance of $0.001$ corresponds to a distance of approximately $111$ meters. 


\subsection{Spatio-Temporal Similarity Results}

Across all approaches regards to spatio-temporal similarity defined in Section~\ref{subsubsec:stsim}, we clearly observe that data dropout has a critical effect on staypoint quality more severe than location noise. However, as that noise becomes severe, we do tend to see significant dropoffs in staypoint quality, especially from a noise standard deviation of 25m to 50m. One notable result is in the single-shot Trackintel baseline: while this approach performs well at low levels of noise, it sees some of the most severe dropoffs in quality as noise level increases. This result is significant, as the Trackintel baseline (an implementation of the algorithm proposed in~\cite{li2008mining}) is the state-of-the-art.
Comparatively, the parameter-optimized Trackintel Hyperband Search does not see nearly so severe a dropoff in quality. However, this parameter-optimized version of the Trackintel baseline is extremely sensitive to its parameters, and will return nearly useless results without these parameters being properly set (as seen in the additional experiments on our Github Repository using different parameters spatial and temporal thresholds). However, in the real-world, it is difficult to infer the exact noise parameters without labelled ground truth of staypoints for supervision. We also note that Trackintel seems more sensitive to data dropout than other approaches, both in the single-shot baseline and in its parameter-optimized version. Also notable among our results are the SSPE and 3-Step HMM-GEM results. These approaches seem more robust to noise and dropout than others, even in the single-shot case- meaning they are also less sensitive to parameters than the other approaches. Finally, we note that at low noise levels, T-DBSCAN does not perform as well as competitors, but sesms comparatively robust to increased noise levels.

\vspace{-0.2cm}
\subsection{Temporal Similarity Results}\vspace{-0.1cm}
We next investigate the performance of all approaches through the lens of temporal similarity defined in Section~\ref{subsubsec:temporalsim}. In this regard, SSPE appears to be the most robust algorithm to noise and data dropout, both in the single-shot and parameter-optimized cases. We also note that the Hybrid Stay Window algorithm is among the most competitive in this metric, especially at lower noise levels, while 3-Step HMM-GEM is more competitive at increased noise levels. Once again, we note the significant performance dropoff for the Trackintel baseline at higher noise levels, showing its dependence on proper parameter settings. In the parameter-optimized case, this dropoff is not as severe, but still tends to be worse than its competitors. In general, dropout level tends to have a much lower impact on this metric than F1-score. This makes sense, since data dropout may remove the start or end time of a staypoint, making the identified staypoint incorrect in terms of F1, while Temporal IoU is measured over all observations in a trajectory. 

\vspace{-0.2cm}
\subsection{Spatial Similarity Results}\vspace{-0.1cm}
We now present the evaluation results based on the spatial similarity described in Section~\ref{subsubsec:spatialsim}. Similar to Temporal IoU, dropout level does not play as significant a role in the results as noise level does. In general, SSPE and 3-Step HMM-GEM appear to be the most robust algorithms to noise when evaluating for Spatial Overlap score. Hybrid Stay Window is also competitive at high noise levels, especially in the parameter-optimized case. Interestingly, with a noise level of 10m, almost all algorithms are competitive, and it becomes more difficult to distinguish between their performances.

\subsection{Runtime Analysis}
We provide a brief review of runtimes for the various algorithms discussed in this paper. For the purposes of this comparison, all algorithms were run on the same machine and environment, using their default parameter settings, to calculate predicted staypoints for a single agent in one dataset. We used the PoL dataset with a noise level of 25 and dropout level of 2 for this purpose, and the same agent was used across all runs. Results can be seen in Table~\ref{tab:runtime}. T-DBSCAN is significantly faster than other algorithms here, while the Centroid Sliding Window algorithm stands out as particularly slow. At the dataset level, most staypoint algorithms should be completely parallelizable by its agents. Notably, we do not include runtimes for parameter-optimized versions, since these are dependent on the range of parameters observed and method for doing so. We do note that the Trackintel Hyperband Search, as described in this paper, took approximately 7 days and 10 hours to complete for all 16 datasets, run in sequence.

\begin{table}
\begin{center}
\begin{tabular}{ |c|c| }
\hline 
Algorithm & Runtime (s) \\ 
\hline

T-DBSCAN & 0.447286 \\ 

SSPE & 2.738956 \\ 

3-Step HMM-GEM & 3.811537 \\ 

Hybrid Stay Window & 4.047646 \\ 

Trackintel Baseline & 2.909153 \\ 

Trackintel w/Filter & 2.917468 \\ 

Centroid Sliding Window & 28.408239 \\

Adaptive radius Sliding Window & 3.057460 \\ 

Histogram Gradient Boosting & 2.086561 \\ 
\hline
\end{tabular}
\captionof{table}{Runtime results to calculate staypoints on a single agent using default parameters}\label{tab:runtime}\vspace{-0.7cm}
\end{center}
\end{table}

\section{Conclusion and Future Work}
\label{sec:conclusion}
The results in this paper show that staypoint detection is still an open problem. The state-of-the-art algorithms struggle under the noise and dropout that we expect to see in real-world data, the negative impacts of which on downstream tasks could be severe. By providing datasets with individual-level trajectories and ground-truth staypoints, we allow for experimental assessment of staypoint detection algorithms, both without knowledge of the ground-truth and under ideal parameter settings. We provide different metrics for a comprehensive assessment of the staypoint detection algorithms. Some novel algorithms proposed in this paper already see improved results from the state-of-the-art, either by being more robust to noise or less dependent on proper parameter settings. These algorithms may be improved in future works, especially in regards to handling data dropout effects, and by testing their robustness to variation in sampling rates.

\section{Acknowledgments}
Supported by the Intelligence Advanced Research Projects Activity (IARPA) via Department of Interior/ Interior Business Center (DOI/IBC) contract number 140D0423C0025. The U.S. Government is authorized to reproduce and distribute reprints for Governmental purposes notwithstanding any copyright annotation thereon. Disclaimer: The views and conclusions contained herein are those of the authors and should not be interpreted as necessarily representing the official policies or endorsements, either expressed or implied, of IARPA, DOI/IBC, or the U.S. Government.

\bibliographystyle{ACM-Reference-Format}
\bibliography{refs/main}

\clearpage
\appendix
\section{Appendix}
\label{sec:appendix}

\subsection{Algorithm Details}
\label{app:algorithm_details}
In this section, the details of three algorithms are provided. Algorithm~\ref{alg:hsw} presents the Hybrid Stay Window (HSW) approach, which scans each individual trajectory using spatial and temporal thresholds while allowing a limited number of noisy observations. Algorithm~\ref{alg:tdbscan} describes the temporal DBSCAN post-processing step, where clusters are split or removed based on temporal gaps and minimum-duration requirements. Algorithm~\ref{alg:sspe} presents the Sequential Stay Point Extraction (SSPE) method, which incrementally builds candidate stay sequences using a dynamic centroid, temporal dropout constraints, and recursive lookahead to distinguish short anomalies from true departures.

\begin{algorithm}[h]
\caption{Hybrid Stay Window (HSW)}
\label{alg:hsw}
\begin{algorithmic}[1]

\fontsize{8pt}{11pt}\selectfont
\setlength{\baselineskip}{9.5pt}

\State \textbf{Input:} Trajectory dataset $T$ containing individual identifiers, locations, and timestamps; spatial threshold $D$; temporal threshold $\Delta t$; noise tolerance parameter $\eta$
\State \textbf{Output:} Set of detected stay points $S$

\State Clean $T$ by removing incomplete records
\State Sort all records by individual identifier and timestamp
\State Initialize $S \gets \emptyset$

\For{each individual trajectory $\tau$ in $T$}
\State Sort $\tau$ chronologically
\State Set starting observation index $i \gets 1$
\While{$i$ is not the final observation in $\tau$}
    \State Set window end index $j \gets i + 1$
    \State Set remaining noise allowance $c \gets \eta$

    \While{$j$ is within $\tau$}
        \State Compute the geographic distance between observations $i$ and $j$

        \If{distance exceeds $D$}
            \If{$c > 0$}
                \State Set $c \gets c - 1$
                \State Set $j \gets j + 1$
                \State Continue
            \EndIf

            \State Define the candidate stay window from observation $i$ to $j-1$
            \State Compute the duration of the candidate stay window

            \If{duration is at least $\Delta t$}
                \State Create a stay point using the mean location of the candidate stay window
                \State Record arrival time, departure time, duration, number of observations, and spatial radius
                \State Add the stay point to $S$
                \State Set $i \gets j$
            \Else
                \State Set $i \gets i + 1$
            \EndIf

            \State Break
        \EndIf

        \State Set $j \gets j + 1$
    \EndWhile

    \If{the end of $\tau$ is reached without finding an exit from the candidate window}
        \State Break
    \EndIf
\EndWhile

\EndFor

\State \Return $S$
\end{algorithmic}
\end{algorithm}

\begin{algorithm}[t]
\caption{Temporal DBSCAN Post-Processing} \label{alg:tdbscan}
\begin{algorithmic}[1]

\fontsize{8pt}{11pt}\selectfont
\setlength{\baselineskip}{8.5pt}
\For{$cluster$ in $clusters$}
    \For{$gap$ in $cluster$}
        \If{len($gap$) $\geq$ $minTime$}
            \State Split $cluster$ into $cluster_1$ and $cluster_2$
            \State Add $cluster_1, cluster_2$ to $clusters$
            \State Delete $cluster$ from $clusters$
            \State \textbf{break}
        \Else
            \State Add $gap$ to $cluster$
        \EndIf
    \EndFor
    \If{len$(cluster) < minTime$}
        \State Delete $cluster$ from $clusters$
    \EndIf
\EndFor
\end{algorithmic}
\end{algorithm}

\begin{algorithm}[t]
\caption{Sequential Stay Point Extraction (SSPE)}
\label{alg:sspe}
\begin{algorithmic}[1]

\fontsize{8pt}{11pt}\selectfont
\setlength{\baselineskip}{8.5pt}

\Statex \textbf{Input:} Trajectory $\mathcal{T} = \{p_1, p_2, \dots, p_N\}$, where $p = (uid, lat, lon, t)$
\Statex \textbf{Parameters:} $\delta_{dist}$ (Spatial threshold), $\tau_{min}$ (Min duration), $\tau_{max}$ (Max dropout), $n$ (Recursive lookahead step)
\Statex \textbf{Output:} Set of Stay Points $\mathcal{S}$
\Statex
\State Initialize $\mathcal{S} \leftarrow \emptyset$, and pointer $i \leftarrow 1$
\While{$i < N$}
    \State Initialize candidate sequence $\mathcal{C} \leftarrow \{p_i\}$
    \State Set evaluation index $j \leftarrow i + 1$
    \While{$j \le N$}
        \State Compute dynamic centroid $\mu_{\mathcal{C}}$ as the mean coordinate of points in $\mathcal{C}$
        \State Calculate temporal gap $\Delta t \leftarrow t_j - t_{j-1}$
        \State Calculate spatial displacement $d \leftarrow \text{Haversine}(\mu_{\mathcal{C}}, p_j)$
        
        \If{$\Delta t > \tau_{max}$}
            \State \textbf{break} \Comment{Signal dropout exceeds tolerance}
        \EndIf
        
        \If{$d \le \delta_{dist}$}
            \State Append $p_j$ to $\mathcal{C}$
            \State $j \leftarrow j + 1$
        \Else \Comment{Distance threshold breached}
            \State Initialize $is\_noise \leftarrow \text{False}$
            \For{$k$ from $j+1$ to $\min(j+n, N)$} \Comment{Recursive lookahead}
                \State Calculate $d_{future} \leftarrow \text{Haversine}(\mu_{\mathcal{C}}, p_k)$
                \State Calculate $\Delta t_{future} \leftarrow t_k - t_{j-1}$
                \If{$d_{future} \le \delta_{dist}$ and $\Delta t_{future} \le \tau_{max}$}
                    \State $is\_noise \leftarrow \text{True}$
                    \State \textbf{break}
                \EndIf
            \EndFor
            
            \If{$is\_noise$}
                \State $j \leftarrow j + 1$ \Comment{Classify $p_j$ as a anomaly and continue}
            \Else
                \State \textbf{break} \Comment{Spatial departure detected}
            \EndIf
        \EndIf
    \EndWhile
    
    \State Evaluate Duration $D \leftarrow t_{\text{last}} - t_{\text{first}}$ for $p \in \mathcal{C}$
    \If{$D \ge \tau_{min}$}
        \State Construct agent level stay point $s = (uid, \mu_{\mathcal{C}}, t_{\text{first}}, t_{\text{last}} )$
        \State $\mathcal{S} \leftarrow \mathcal{S} \cup \{s\}$
        \State Update $i \leftarrow \text{index}(p_{\text{last}}) + 1$
    \Else
        \State $i \leftarrow i + 1$
    \EndIf
\EndWhile

\State \Return $\mathcal{S}$
\end{algorithmic}
\end{algorithm}

\break
\subsection{Additional Experimental Results}\label{app:results}

\begin{table*}[h]
\centering
\caption{Recall-score results across the PoL dataset with different noise and dropout levels from various methods}
\label{tab:recall-results}
\small
\setlength{\tabcolsep}{3pt}
\resizebox{\textwidth}{!}{
\begin{tabular}{c cccc cccc cccc cccc}
\toprule

Noise level
& \multicolumn{4}{c}{0m} 
& \multicolumn{4}{c}{10m} 
& \multicolumn{4}{c}{25m} 
& \multicolumn{4}{c}{50m} \\

\cmidrule(lr){2-5} \cmidrule(lr){6-9} \cmidrule(lr){10-13} \cmidrule(lr){14-17}

Dropout Level
& 0 & 1 & 2 & 3
& 0 & 1 & 2 & 3
& 0 & 1 & 2 & 3
& 0 & 1 & 2 & 3 \\

\midrule

\multicolumn{17}{c}{\textbf{Single-Shot (Unsupervised) Approaches}} \\

\midrule

T-DBSCAN
& 0.944135 & 0.880710 & 0.705069 & 0.430616 
& 0.938047 & 0.873712 & 0.696689 & 0.418730 
& 0.904832 & 0.839555 & 0.672901 & 0.399481 
& 0.743248 & 0.675284 & 0.519363 & 0.296874 \\

SSPE
& \textbf{0.9906} & \textbf{0.9271} & \underline{\textbf{0.7620}} & \underline{\textbf{0.4825}} 
& \underline{\textbf{0.9853}} & \underline{\textbf{0.9200}} & \underline{\textbf{0.7561}} & \underline{\textbf{0.4788}} 
& \underline{\textbf{0.9780}} & \underline{\textbf{0.9134}} & \underline{\textbf{0.7560}} & \underline{\textbf{0.4764}} 
& \underline{\textbf{0.9059}} & \underline{\textbf{0.8471}} & \underline{\textbf{0.6991}} & \underline{\textbf{0.4456}} \\

3-Step HMM-GEM
& \underline{\textbf{0.993776}} & \underline{\textbf{0.927755}} & \textbf{0.761678} & \textbf{0.480804} & \textbf{0.972584} & \textbf{0.906441} & \textbf{0.746391} & \textbf{0.471179} & \textbf{0.945061} & \textbf{0.881606} & \textbf{0.733017} & \textbf{0.461874} & \textbf{0.865818} & \textbf{0.809543} & \textbf{0.671778} & \textbf{0.428522} \\

Hybrid Stay Window
 & 0.9639  & 0.9024  & 0.7438  & 0.4712 
 & 0.9634  & 0.9002  & 0.7418  & 0.4689 
 & 0.8434  & 0.7918  & 0.6623  & 0.4246 
 & 0.1579  & 0.1506  & 0.1315  & 0.0935  \\

Trackintel Baseline
& 0.926890 & 0.780472 & 0.592944 & 0.374706 
& 0.922927 & 0.774839 & 0.588359 & 0.369575 
& 0.133560 & 0.127017 & 0.112003 & 0.078500 
& 0.001199 & 0.001199 & 0.001154 & 0.000744 \\

Trackintel w/Filter
& 0.926890 & 0.780092 & 0.592716 & 0.374584 
& 0.922927 & 0.774414 & 0.588025 & 0.369423 
& 0.133560 & 0.126941 & 0.111852 & 0.078363 
& 0.001199 & 0.001078 & 0.001002 & 0.000683 \\

Centroid Sliding Window
  & 0.9333 & 0.8704 & 0.7051 & 0.4433
  & 0.9335 & 0.8675 & 0.7123 & 0.4511
  & 0.7230 & 0.6790 & 0.5715 & 0.3631
  & 0.0366 & 0.0350 & 0.0316 & 0.0222 \\
\midrule

\multicolumn{17}{c}{\textbf{Parameter-Optimized Approaches}} \\

\midrule

T-DBSCAN (Grid Search)
& 0.957464 & 0.886205 & 0.716591 & 0.440712 
& 0.942374 & 0.874865 & 0.709729 & 0.431861 
& 0.906441 & 0.844109 & 0.689751 & 0.417303 
& \textbf{0.805717} & \textbf{0.747226} & 0.598288 & 0.353013 \\

SSPE  (Grid Search)
& \underline{\textbf{0.9936}} & \underline{\textbf{0.9287}} & \underline{\textbf{0.7651}} & \underline{\textbf{0.4826}}
& \underline{\textbf{0.9866}} & \underline{\textbf{0.9201}} & \underline{\textbf{0.7561}} & \underline{\textbf{0.4788}} 
& \underline{\textbf{0.9780}} & \underline{\textbf{0.9134}} & \underline{\textbf{0.7561}} & \underline{\textbf{0.4765}}
& \underline{\textbf{0.9196}} & \underline{\textbf{0.8591}} & \underline{\textbf{0.7074}} & \underline{\textbf{0.4485}} \\

Trackintel Hyperband Search
& \textbf{0.9796} & 0.9089 & 0.7195 & 0.4340
& \textbf{0.9752} & 0.8964 & 0.7092 & 0.4316
& \textbf{0.9416} & \textbf{0.8717} & \textbf{0.6945} & 0.4258
& 0.7740 & 0.7238 & 0.5893 & 0.3678 \\

Hybrid Stay Window
& 0.9787  & \textbf{0.9165}  & \textbf{0.7557}  & \textbf{0.4779} 
& 0.9634  & 0.8975  & 0.7391  & 0.4671 
& 0.8862  & 0.8281  & 0.6884  & \textbf{0.4337} 
& 0.7898  & 0.7397  & \textbf{0.6146}  & \textbf{0.4047} \\

Adaptive-radius Sliding Window
& 0.9738 & 0.9117 & 0.7521 & 0.4754
& 0.9675 & \textbf{0.9041} & \textbf{0.7454} & \textbf{0.4719} 
& 0.6528 & 0.6070 & 0.5076 & 0.3193 
& 0.6321 & 0.5920 & 0.4868 & 0.3178 \\

\midrule

\multicolumn{17}{c}{\textbf{Fully Supervised Approaches}} \\

\midrule

Histogram Gradient Boosting
& 0.9962 & 0.9271 & 0.7498 & 0.4778
& 0.9700 & 0.9130 & 0.7382 & 0.4724
& 0.9759 & 0.9076 & 0.7409 & 0.4119
& 0.9438 & 0.8721 & 0.7187 & 0.4413 \\
\bottomrule
\end{tabular}
}

\end{table*}

\begin{table*}[ht]
\centering
\caption{Precision-score results across the PoL dataset with different noise and dropout levels from various methods}
\label{tab:precision-results}
\small
\setlength{\tabcolsep}{3pt}
\resizebox{\textwidth}{!}{
\begin{tabular}{c cccc cccc cccc cccc}
\toprule

Noise level
& \multicolumn{4}{c}{0m} 
& \multicolumn{4}{c}{10m} 
& \multicolumn{4}{c}{25m} 
& \multicolumn{4}{c}{50m} \\

\cmidrule(lr){2-5} \cmidrule(lr){6-9} \cmidrule(lr){10-13} \cmidrule(lr){14-17}

Dropout Level
& 0 & 1 & 2 & 3
& 0 & 1 & 2 & 3
& 0 & 1 & 2 & 3
& 0 & 1 & 2 & 3 \\

\midrule

\multicolumn{17}{c}{\textbf{Single-Shot (Unsupervised) Approaches}} \\

\midrule

T-DBSCAN
& 0.976051 & 0.924660 & 0.761556 & 0.540017 
& 0.971819 & 0.919458 & 0.749616 & 0.513007 
& \textbf{0.933593} & 0.874002 & 0.711668 & 0.477982 
& 0.746663 & 0.671952 & 0.504382 & 0.295528 \\

SSPE
& \textbf{0.9980} & \underline{\textbf{0.9535}} & 0.8397 & \textbf{0.6762} 
& \underline{\textbf{0.9951}} & \underline{\textbf{0.9500}} & \underline{\textbf{0.8354}} & \underline{\textbf{0.6757}} 
& \underline{\textbf{0.9890}} & \underline{\textbf{0.9430}} & \underline{\textbf{0.8348}} & \underline{\textbf{0.6683}} 
& \underline{\textbf{0.9029}} & \underline{\textbf{0.8636}} & \underline{\textbf{0.7634}} & \underline{\textbf{0.6159}} \\

3-Step HMM-GEM
& \underline{\textbf{1.0}} & \textbf{0.953298} & \textbf{0.844069} & 0.674647 & 0.969977 & 0.924862 & 0.820681 & 0.660046 & 0.928804 & \textbf{0.885747} & \textbf{0.793956} & \textbf{0.636573} & \textbf{0.80618} & \textbf{0.772441} & \textbf{0.694803} & \textbf{0.56421} \\

Hybrid Stay Window
 & 0.9915  & 0.9488  & \underline{\textbf{0.8445}}  & \underline{\textbf{0.6795}} 
 & \textbf{0.9791}  & \textbf{0.9365}  & \textbf{0.8321}  & \textbf{0.6718} 
 & 0.6615  & 0.6375  & 0.5793  & 0.4804 
 & 0.0143  & 0.0142  & 0.014  & 0.0138  \\

Trackintel Baseline
& 0.974589 & 0.849715 & 0.729461 & 0.641084 
& 0.972067 & 0.844602 & 0.723092 & 0.631567 
& 0.010326 & 0.010235 & 0.010207 & 0.009953 
& 0.000301 & 0.000310 & 0.000338 & 0.000306 \\

Trackintel w/Filter
& 0.974589 & 0.850736 & 0.730682 & 0.641926 
& 0.972067 & 0.845692 & 0.723870 & 0.632160 
& 0.010326 & 0.010241 & 0.010206 & 0.009945 
& 0.000301 & 0.000282 & 0.000298 & 0.000284 \\

 Centroid Sliding Window
  & 0.9528 & 0.9014 & 0.7871 & 0.6237
  & 0.9529 & 0.9020 & 0.7910 & 0.6269
  & 0.5657 & 0.5434 & 0.4970 & 0.4048
  & 0.0023 & 0.0023 & 0.0023 & 0.0023 \\
  
\midrule

\multicolumn{17}{c}{\textbf{Parameter-Optimized Approaches}} \\

\midrule

T-DBSCAN (Grid Search)
& 0.982307 & 0.924140 & 0.793105 & 0.573667 
& 0.969545 & 0.924466 & 0.775772 & 0.556431 
& 0.950500 & 0.904873 & 0.773257 & 0.542054 
& \textbf{0.888234} & \textbf{0.835353} & \textbf{0.696026} & 0.435504 \\

SSPE  (Grid Search)
& \underline{\textbf{0.9967}} & \underline{\textbf{0.9530}} & \underline{\textbf{0.8445}} & 0.6763 
& \underline{\textbf{0.9945}} & \underline{\textbf{0.9500}} & \textbf{0.8354} & 0.6757 
& \underline{\textbf{0.9891}} & \underline{\textbf{0.9431}} & \underline{\textbf{0.8348}} & \textbf{0.6683} 
& \underline{\textbf{0.9409}} & \underline{\textbf{0.8984}} & \underline{\textbf{0.7927}} & \underline{\textbf{0.6349}} \\

Trackintel Hyperband Search
& \textbf{0.9955} & 0.9430 & 0.8303 & \underline{\textbf{0.7158}}
& \textbf{0.9847} & 0.9371 & 0.8277 & \underline{\textbf{0.7042}}
& \textbf{0.9583} & \textbf{0.9074} & \textbf{0.8098} & \underline{\textbf{0.6811}}
& 0.5851 & 0.5593 & 0.4852 & 0.4344 \\

Hybrid Stay Window
 & 0.9954  & \textbf{0.9512}  & \textbf{0.8442}  & \textbf{0.677} 
 & 0.9791  & \textbf{0.9398}  & \underline{\textbf{0.8365}}  & \textbf{0.6777} 
 & 0.9124  & 0.8733  & 0.785  & 0.6307 
 & 0.6709  & 0.6463  & 0.5847  & \textbf{0.4731}  \\

Adaptive-radius Sliding Window
& 0.9936 & 0.9493 & 0.8427 & 0.6757
& 0.9678 & 0.9242 & 0.8209 & 0.6628 
& 0.7281 & 0.6933 & 0.6218 & 0.5007 
& 0.5994 & 0.5754 & 0.5229 & 0.4296 \\

\midrule

\multicolumn{17}{c}{\textbf{Fully Supervised Approaches}} \\

\midrule

 Histogram Gradient Boosting
  & 1.0 & 0.9671 & 0.8907 & 0.7784
  & 0.9727 & 0.9458 & 0.8278 & 0.6943
  & 0.9895 & 0.9366 & 0.8407 & 0.5477
  & 0.9603 & 0.9115 & 0.8306 & 0.6675 \\
\bottomrule
\end{tabular}
}

\end{table*}

\end{document}